\documentclass[journal]{IEEEtran}
\usepackage{multirow}
\usepackage[caption=false]{subfig}

\usepackage{ifpdf}
 \ifpdf
 \else
 \fi

\ifCLASSINFOpdf
   \usepackage[pdftex]{graphicx}
   \graphicspath{{../pdf/}{../jpg/}}
   \DeclareGraphicsExtensions{.pdf,.jpg,.png}
\else
   \usepackage[dvips]{graphicx}
   \graphicspath{{../eps/}}
   \DeclareGraphicsExtensions{.eps}
\fi

\usepackage{amsmath}

\begin{document}

\title{A Learning-based Optimization Algorithm:\\Image Registration Optimizer Network}

\author{Jia~Wang, Ping~Wang, Biao~Li, Yinghui~Gao, and~Siyi~Zhao
\thanks{J. Wang, P. Wang, B. Li and Y. Gao are with the Key Laboratory of ATR, College of Electronic Science and Technology, National University of Defense Technology, Changsha, 410073, China. (e-mail: wangj\_cs04@sina.com; 760521407@qq.com; libiao\_cn@163.edu.cn; yhgao@nudt.edu.cn).}
\thanks{S. Zhao is with Domestic foundation Software Engineering Research Center, College of Computer Science and Technology,National University of Defense Technology, Changsha, 410073, China. (e-mail: 343422652@qq.com)}
}

\markboth{Journal of \LaTeX\ Class Files,~Vol.~14, No.~8, November~2020}%
{Shell \MakeLowercase{\textit{et al.}}: Bare Demo of IEEEtran.cls for IEEE Journals}

\maketitle

\begin{abstract}
Remote sensing image registration is valuable for image-based navigation system despite posing many challenges. As the search space of registration is usually non-convex, the optimization algorithm, which aims to search the best transformation parameters, is a challenging step. Conventional optimization algorithms can hardly reconcile the contradiction of simultaneous rapid convergence and the global optimization. In this paper, a novel learning-based optimization algorithm named Image Registration Optimizer Network (IRON) is proposed, which can predict the global optimum after single iteration. The IRON is trained by a 3D tensor (9×9×9), which consists of similar metric values. The elements of the 3D tensor correspond to the 9×9×9 neighbors of the initial parameters in the search space. Then, the tensor’s label is a vector that points to the global optimal parameters from the initial parameters. Because of the special architecture, the IRON could predict the global optimum directly for any initialization. The experimental results demonstrate that the proposed algorithm performs better than other classical optimization algorithms as it has higher accuracy, lower root of mean square error (RMSE), and more efficiency. Our IRON codes are available for further study\footnote{https://www.github.com/jaxwangkd04/IRON}.
\end{abstract}


\IEEEpeerreviewmaketitle

\section{Introduction}

\IEEEPARstart{I}{mage} registration between sensed and reference images is of vital importance for visual homing systems in remote sensing image applications. The task of image registration is to align the sensed image - which are captured with different modalities, from different viewpoints, or at different times - with the reference image.

Classical image registration algorithms can be generally divided into two categories \cite{Majiayi:A survey}: intensity-based algorithms and feature based algorithms. Intensity-based algorithms directly calculate the similarity metric in intensity, such as mutual information or cross correlation. However, they fail in multi-model images registration because of their non-linear intensity relationships. Instead of utilizing intensity directly, feature-based algorithms attempt to estimate the similarity metric between features such as point, contour, line, section, etc. The commonly used similarity metrics for feature-based algorithms include Minkowski distance, Euclidean distance, Kullback-Leibler (KL) divergence, Hausdorff distance, and others. Both intensity- and feature-based algorithms align the image pair by optimizing transformation parameters. Therefore, optimal transformation parameters indicate the completeness of the registration. Once the image feature and similarity metric are formulated, the rest of image registration can be treated as an optimization problem.

The focus of this paper is to find the optimal parameters. Traditional optimization algorithms are from a basic perspective divided into three categories: 1) first-order derivative methods with the representative of gradient descent \cite{Bottou:SGD} and its variants; 2) high-order derivative methods, such as Newton’s method \cite{Kelley:NEWTON} as well as its variants ; 3) heuristic derivative-free methods, among which genetic algorithm \cite{Whitley:GA} and simulated annealing algorithm \cite{Van:simulatedAnnealing} are typical. First order derivative methods iteratively update variables in the opposite direction of the gradients of the objective function. Coherent point drift (CPD) \cite{Myronenko:CPD} is one of the most representative methods. CPD method treats registration  as maximizing the posterior probability of the point set, and it updates the transformation parameters by solving a first-order differential equation. Unfortunately, the first order derivative method converges slowly and is easily trapped in a  local minimum, especially in the case of badly scaled or severely degraded data. The high-order derivative methods converge faster because a quadric surface would fit the search space better than a linear plane. Chen et al. \cite{Chen:two-GaussNewton} proposed a two-step Gauss-Newton method; the second step is to minimize a quadratic approximation of the objective function. Experimental results showed that it outperformed the standard Gauss-Newton method. However, the calculation of the inverse Hessian matrix is expensive; the quasi-Newton algorithm address this shortcoming by using an approximation. Min et al. \cite{Min:quasi-newton} adopted a coarse-to-fine strategy based on a quasi-Newton algorithm to determine the optimal transformation parameters. It achieved more accurate registration in a shorter runtime. When the derivative of the objective function does not exist, the heuristic derivative-free method was proposed. Inspired by ant colonies’ foraging behavior, Wu et al. \cite{Wu:ant colony} combined a continuous ant colony optimization algorithm with an efficient local search operation. It improves efficiency and accuracy in multi-sensor remote sensing image registration. Zhao et al. \cite{Zhao:annealing} utilized a dynamic threshold strategy to compute the prior probability between features and adopted the deterministic annealing method to optimize the optimal correspondence. This achieved an optimal mapping from a local to a global scale. Wu et al. \cite{Wu:particle swarm} directly sampled the image registration transformation parameters rather than using the RANSAC method and utilized the particle swarm optimization algorithm to optimize parameters and achieve a better performance than with traditional RANSAC. However, heuristic derivative-free methods cannot guarantee the optimum theoretically and suffer from contradictory rapid convergence and a global optimum.

Deep learning technology has been proven successful in computer vision tasks, and the step of image registration optimization can be treated as typical regression processing. Deep neural network encapsulates image processing steps into a black box, which makes it easy and convenient to design an image registration optimization algorithm. Recently, Rocco et al. \cite{Rocco:geometric matching} fed two images into a “Siamese” architecture to obtain descriptors, and then matched the descriptors by a tentative correspondence map. The regression network outputs geometric transformation parameters through fully connected layers directly. Instead of directly computing the parameters from fully connected layers, Poursaeed et al. \cite{Poursaeed:fundamental} introduced a reconstruction as well as normalization layers after the fully connected layers and achieved better performance. However, the above methods were trained by images - which is image dependent and restricts its application scope. In this paper, an image-independent registration optimization network is proposed. It is an optimizer that can be applied to any image registration optimization step.

All optimization algorithms need to reconcile the “global optimum” and “rapid convergence” simultaneously. Unfortunately, most traditional optimization approaches cannot reconcile these effectively. To alleviate the problem, a learning-based optimization method named the image registration optimization network (IRON) is proposed here. It differs from traditional DNN architecture, its training data is a tensor rather than an image, and the tensor’s label is a vector in the search space. The tensor consists of similarity metric values, and the vector points to the global optimal parameters from the initialized parameters. Therefore, it is expected that, by feeding the special training dataset into the network, the trained IRON will predict the global optimal parameters in a straightforward manner. It implies that the IRON has learned the whole structure information of the search space.

The rest of this paper is organized as follows: Section II describes the proposed approach in detail; Section III presents experiments and analysis; and conclusions are presented in Section IV.

\section{Proposed Approach}
\subsection{Motivation}
\begin{figure}[htpb]
	\centering
	\subfloat[a better searching result]{\includegraphics[width=0.45\linewidth]{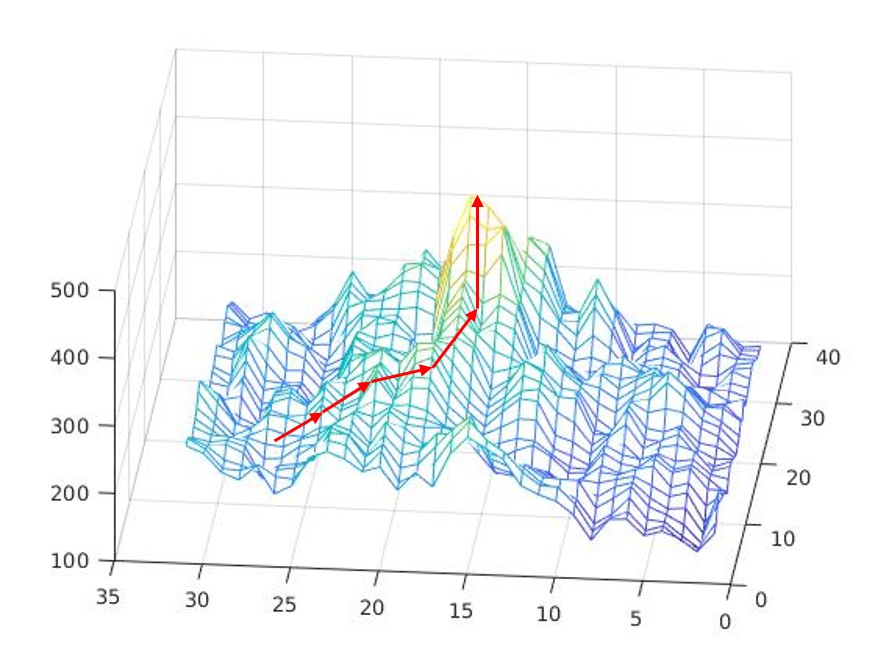}}
	\label{subfig:p1}
	\hspace{1em}
	\subfloat[full searching result]{\includegraphics[width=0.45\linewidth]{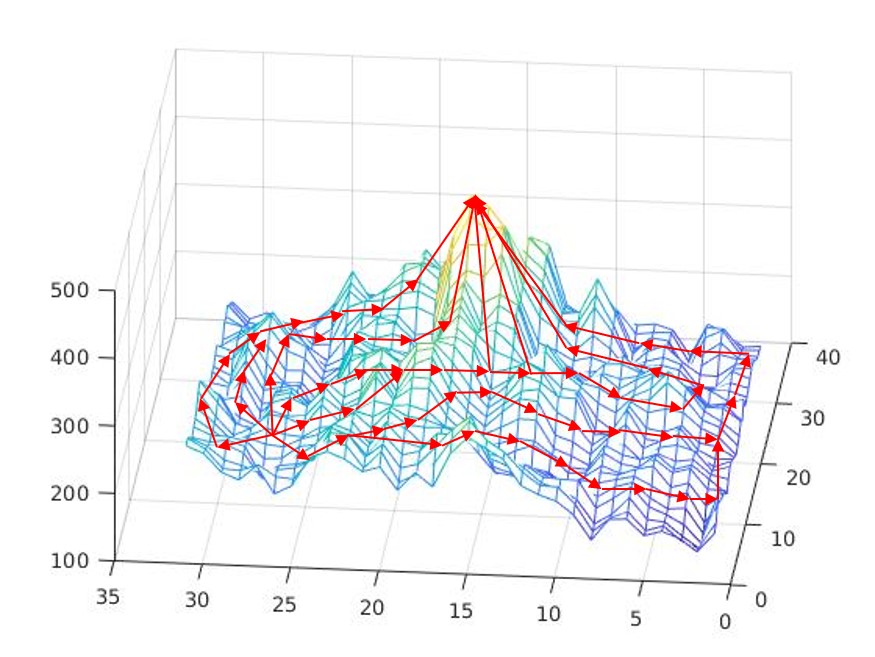}}
	\label{subfig:p3}
	
	\subfloat[local maximum result]{\includegraphics[width=0.45\linewidth]{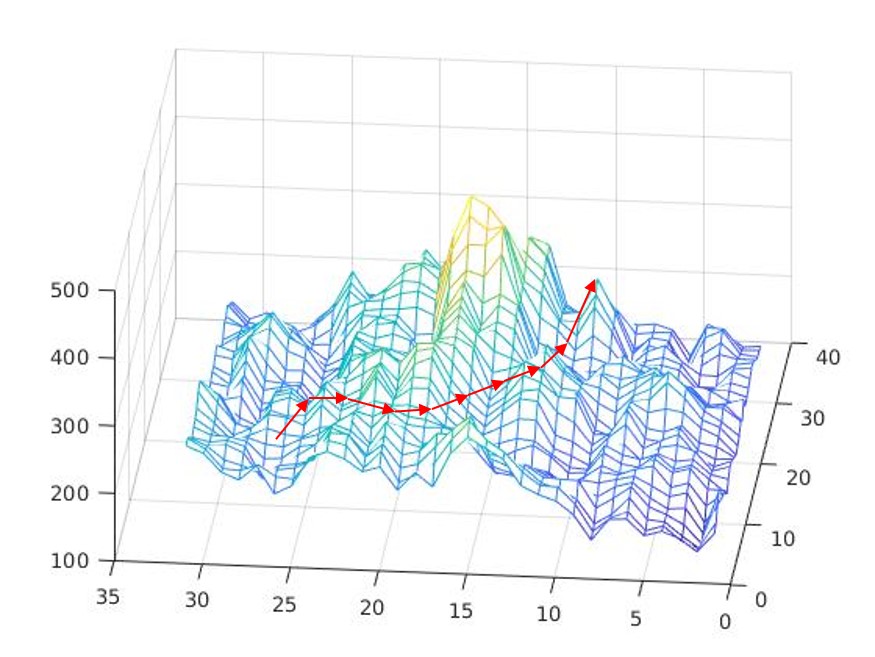}}
	\label{subfig:p2}
	\hspace{1em}
	\subfloat[purposed searching result]{\includegraphics[width=0.45\linewidth]{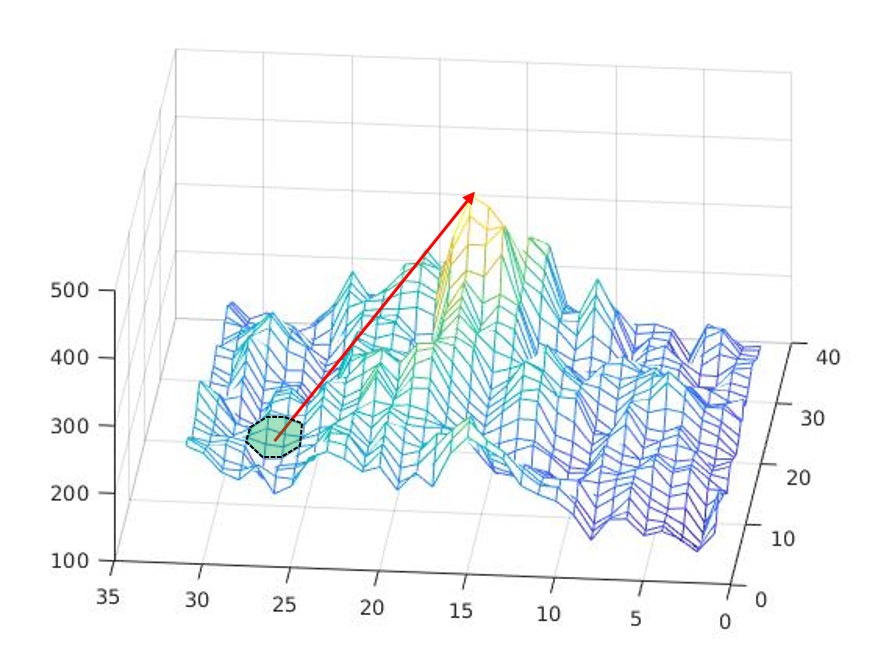}}
	\label{subfig:p4}
	
	\caption{Optimization steps.}
	\label{fig:diff_searching_path}
\end{figure}

We hope to reconcile the common contradiction of global optimization and rapid convergence in optimization methods by deep learning-based technology. First, let us depict the contradiction of the optimization algorithm and visually indicate our purpose. Assuming randomly initialized parameters in the search space, a better optimization algorithm is expected to search the optimal parameters as depicted by the arrow in Fig.\ref{fig:diff_searching_path}(a). However, most search spaces are usually non-convex and oscillating. Consequently, most optimization algorithms cannot reach a global optimal solution in several searching steps. Thus, we have two results. Fig.\ref{fig:diff_searching_path}(b) shows that the optimization algorithm has climbed up to a local maximum after a few steps, and Fig.\ref{fig:diff_searching_path}(c) depicts that the optimum has been found by searching almost every corner of the search space. Obviously, neither of two above situations meet our expectations. Our goal is to overcome these two shortcomings in an optimization algorithm.

Deep learning has reached excellent performance in most computer vision tasks by training with labeled images. Inspired by this, we deliberated whether  a labeled similarity metric tensor could be used to train a deep neural network. If the network learned the structure of the search space, for a random initialization, it could predict the optimum directly rather than seeking iteratively or densely. Our purposing is depicted in Fig.\ref{fig:diff_searching_path}(d). Next, we introduce how to generate a similarity metric tensor and its label, then describe the details of the image registration optimization network which can be applied as an optimizer in any registration algorithm.

\subsection{Image registration optimization network}
\begin{figure}[htpb]
	\centering
	\subfloat[$\tau$]{\includegraphics[width=0.45\linewidth]{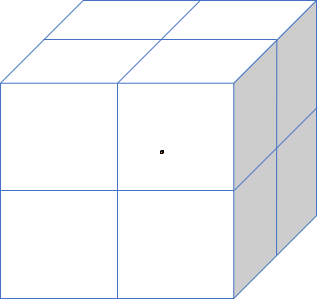}}
	\label{subfig:whole}
	\hspace{1em}
	\subfloat[$\tau'$ in $\tau$]{\includegraphics[width=0.45\linewidth]{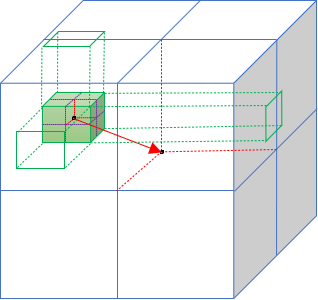}}
	\label{subfig:whole-sub-tensor}
	
	\caption{Similarity metric tensor.}
	\label{fig:similarity_metric_tensor}
\end{figure}

Throughout this paper, we use the following notations:

$I_s ,I_t :R^2  \to R$, - the sensed image and reference image,

$M,N\in R$, - the number of feature points from a sensed and reference image,

${\bf{U}} = ({\bf{u}}_1 ,...,{\bf{u}}_M )^T  \in R^{M \times 2} $, - the sensed image point set,

$\mathbf{V}=\left( \mathbf{v}_1,...,\mathbf{v}_N \right) ^T\in R^{N\times 2}$, - the reference image point set,

${\bf{{\rm T}}}{\rm{(}}{\bf{U}},{\bf{\theta }})$, - transformation ${\bf{{\rm T}}}$ applied to ${\bf{U}}$, where ${\bf{\theta }}$ is a set of transformation parameters to be optimized,
in this paper $\theta \in R^3$,

${\bf{\tau }} = [\tau _{ijk} ] \in R^{S \times S \times S}$, - the whole search space similarity metric tensor which is shown in Fig.\ref{fig:similarity_metric_tensor}(a), $\tau _{ijk}  = O\left( {{\bf{{\rm \bf{T}}}}{\rm{(}}{\bf{U}},\theta _{x_i } ,\theta _{y_j } ,\theta _{z_k }, \alpha ,\beta ,\gamma ),{\bf{V}}} \right)$ 
and $i,j,k=1,...,S$, here $S=31$, the tensor was shown in Fig.\ref{fig:similarity_metric_tensor}(b) and the black-dot is global optimum parameters,

${\bf{\tau '}} = [\tau _{lmn} ] \in R^{(b + 1) \times (b + 1) \times (b + 1)}$, - the training data tensor which is the green sub-tensor shown in Fig.\ref{fig:similarity_metric_tensor}(b), and
$l = i - {b \mathord{\left/
		{\vphantom {b 2}} \right.
		\kern-\nulldelimiterspace} 2},...,i + {b \mathord{\left/
		{\vphantom {b 2}} \right.
		\kern-\nulldelimiterspace} 2}$,
$m = j - {b \mathord{\left/
		{\vphantom {b 2}} \right.
		\kern-\nulldelimiterspace} 2},...,j + {b \mathord{\left/
		{\vphantom {b 2}} \right.
		\kern-\nulldelimiterspace} 2}$,
$n = k - {b \mathord{\left/
		{\vphantom {b 2}} \right.
		\kern-\nulldelimiterspace} 2},...,k + {b \mathord{\left/
		{\vphantom {b 2}} \right.
		\kern-\nulldelimiterspace} 2}$,
here $b=8$,

$\iota' \in R^{3} $, - the label of $\tau'$ depicted by the red arrow shown in Fig.\ref{fig:similarity_metric_tensor}(b); it starts from the center of the sub-tensor to the global optimum.

\begin{figure}[htpb]
	\centering
	\includegraphics[width=2.5in]{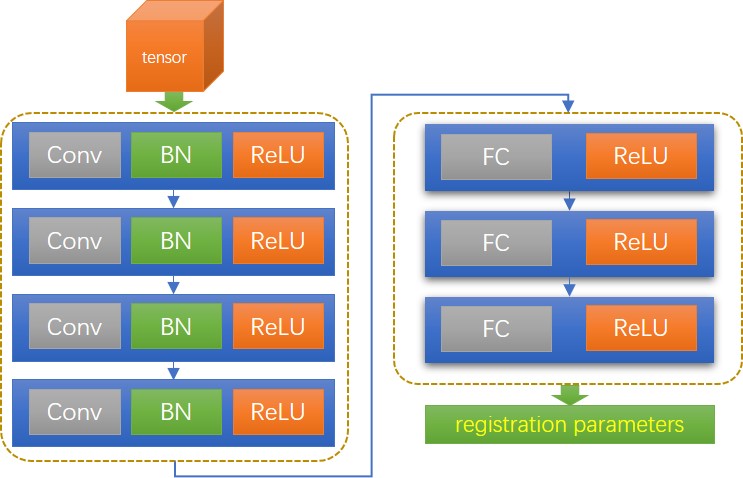}
	\caption{Architecture of IRON. Layer types: Conv: convolution, BN: batch-normalization, FC: fully-connected. We don’t pad the convolution, because the similarity metric is of different from the image density, and pooling layers are not adopted due to the input tensor’s width and height is small. The shape of input tensor is 9×9×9. For CNN-module layers, each layers input channel and output channel is: 1,64; 64,128; 128,256; 256,512; All CNN-module layers use 3×3×3 kernels. For FC layers, each layers I/O channel are: 512,256; 256,64; 64,16; 16,6. All Conv and FC layers use ReLU activation except for the last FC.}
	\label{networkArchitecture}
\end{figure}

The task of the image registration optimization algorithm is to find the optimal parameters. It is described mathematically by the following formula:
$${\bf{\theta }}^ *   = \mathop {\arg }\limits_{\bf{\theta }} \min \left\{ {O\left( {{\bf{{\rm T}}}{\rm{(}}{\bf{U}},{\bf{\theta }}),{\bf{V}}} \right) + \lambda \left\| {\bf{T}} \right\|} \right\}$$
Where $O$ is the similarity function, and $\bf{\lambda }$ a weight that controls the tradeoff between the two terms. For any initialization parameters $\left( {\theta _{x_i } ,\theta _{y_j } ,\theta _{z_k } } \right)$,
the IRON’s output 
$\left( {\hat \theta _x^ *  {\rm{ - }}\theta _{x_i } ,\hat \theta _y^ *  {\rm{ - }}\theta _{y_j } ,\hat \theta _z^ *  {\rm{ - }}\theta _{z_k } } \right)$
plus $\left( {\theta _{x_i } ,\theta _{y_j } ,\theta _{z_k } } \right)$, 
then the result will be the estimation of optimal parameters. The architecture of our proposed IRON is demonstrated in Fig.\ref{networkArchitecture}. In this architecture, the first key step is generating the labeled training datasets.

\subsubsection{Similarity metric tensor generation}
Assuming that there exists a perspective distortion between the sensed image $I_s$ and the reference image $I_t$, the perspective transformation contains three translation parameters (north: $\theta_x$, altitude: $\theta_y$, east: $\theta_z$) and three rotation angle parameters (yaw: $\theta_\alpha$, pitch: $\theta_\beta$, roll: $\theta_\gamma$). Since the three rotation angle parameters 
$\left( {\theta _\alpha  ,\theta _\beta  ,\theta _\gamma  } \right)$ are usually accurate, here we use $\left({\alpha,\beta,\gamma}\right)$ to represent them, and there remain three translation parameters $\left( {\theta _x ,\theta _y ,\theta _z } \right)$ to be optimized. 
 
The search space of image registration in perspective distortion case is not infinite, and the range of each translation parameter could be represented as $\theta _{x_1 }  \le \theta _x  \le \theta _{x_S }$, $\theta _{y_1 }  \le \theta _y  \le \theta _{y_S }$, $\theta _{z_1 }  \le \theta _z  \le \theta _{z_S }$.
We can calculate the whole search space similarity metric tensor using the following steps:

Step 1:  we extract point sets from both images with the SIFT descriptor (for a uni-modal  image) or the Sobel detector (for a multi-modal  image);

Step 2: we execute a perspective transformation on the sensed image point set with parameter $\left( {\theta _{x_i } ,\theta _{y_j } ,\theta _{z_k } ,\alpha ,\beta ,\gamma } \right)$;

Step 3: we calculate the similarity metric between the point sets: $\tau _{ijk}  = O\left( {{\bf{{\rm \bf{T}}}}{\rm{(}}{\bf{U}},\theta _{x_i } ,\theta _{y_j } ,\theta _{z_k }, \alpha ,\beta ,\gamma ),{\bf{V}}} \right)$. Here we have adopted the correlation of the Gaussian Mixture Model as a similarity metric.

We repeat Step 2 and Step 3 for all possible transformation parameters. Then we will obtain a 3D tensor ${\bf{\tau }} = \left[ {\tau _{ijk} } \right] \in R^{S \times S \times S} $, which is the so-called whole search space similarity metric 3D tensor. It is noteworthy that when some angle parameters are inaccurate, we can solve the registration in a similar way by generating a 4D, 5D, or 6D tensor.

Repeat step 2 and 3 for all possible transformation parameters, then we will get a 3D tensor ${\bf{\tau }} = \left[ {\tau _{ijk} } \right] \in R^{S \times S \times S} $, which is the so-called whole search space similarity metric 3D tensor. It is noteworthy that when some angle parameters are inaccurate, we can solve the registration in a similar way by generating a 4D, 5D or 6D tensor.

\subsubsection{Training data generation}
The training data is a sub-tensor ${\bf{\tau '}}$, which is cut from ${\bf{\tau}}$. Supposing that the optimal parameters are $\left( {\theta _x^ *  ,\theta _y^ *  ,\theta _z^ *  } \right)$, for any initial parameters $\left( {\theta _{x_i } ,\theta _{y_j } ,\theta _{z_k } } \right)$,
we cut off a tensor ${\bf{\tau '}} = \left[ {\tau _{lmn} } \right]$ from ${\bf{\tau}}$, where
$l = i - {b \mathord{\left/
		{\vphantom {b 2}} \right.
		\kern-\nulldelimiterspace} 2},...,i + {b \mathord{\left/
		{\vphantom {b 2}} \right.
		\kern-\nulldelimiterspace} 2}$,
$m = j - {b \mathord{\left/
		{\vphantom {b 2}} \right.
		\kern-\nulldelimiterspace} 2},...,j + {b \mathord{\left/
		{\vphantom {b 2}} \right.
		\kern-\nulldelimiterspace} 2}$,
$n = k - {b \mathord{\left/
		{\vphantom {b 2}} \right.
		\kern-\nulldelimiterspace} 2},...,k + {b \mathord{\left/
		{\vphantom {b 2}} \right.
		\kern-\nulldelimiterspace} 2}$.
Thus, the label of ${\bf{\tau '}}$  is defined as: ${\bf{\iota '}} = \left( {\theta _x^ *  {\rm{ - }}\theta _{x_i } ,\theta _y^ *  {\rm{ - }}\theta _{y_j } ,\theta _z^ *  {\rm{ - }}\theta _{z_k } } \right)$, which is a vector that points to global optimal parameters from initial parameters.

\subsubsection{Network details}
In Fig.\ref{networkArchitecture} we see that the proposed image registration optimization network consists of convolution layers (Conv), batch normalization layers (BN), rectified linear units (ReLU), and full connection layers (FC). One could treat it as an optimizer, a predictor, or a regressor. It is worth noting that our IRON is different from other optimization methods such as: SGD, Adam, or RMSProp. These are adopted to train a neural network. Here we adopted Adam to train our IRON. The learning rate was 0.001, and the loss function we employed was mean-square-error. Our implementation is based on pytorch and fastai . The training of each epoch takes about 5 minutes on one Quadro P5000 GPU, Intel Xeon 2.5GHz, 16G RAM. The network is trained with 40 epochs, and it converges before the end of the epochs.

\section{Experiments and Discussion}
\subsection{Data Description}
For the sake of practicability, at the training phase, the sensed images were extracted from satellite images (the spatial resolution is 1.0 m/pixel from Google Maps) through a series of navigation orbit parameters. At the testing phase, the sensed images were real ground images which were acquired by unmanned aerial vehicle (optical images) or airplane (infrared images) at the same scene. In order to get enough training data, we extended one navigation orbit to a cluster of orbits by rotating the orbit’s orientation by 30 degree. This ensured that all possible orbits and the corresponding sensed images were included in the training dataset and avoided under-fitting of the network. All test images were acquired from Guangxi Province, China.

\subsection{Experimental Settings and Evaluation Criteria}
In order to quantitatively validate the effectiveness of the proposed algorithm, our approach was compared with four classical heuristic derivative-free optimization algorithms: pattern search (PS), simulated annealing (Anneal), the genetic algorithm (GA), and particle swam optimization (PSO). All the algorithms were applied as off-the-shelf functions in MATLAB. In order to demonstrate the general applicability of our approach, we tested uni-modal and multi-modal  image registration respectively. The performance index we used was: 1) parameters accuracy (ParamAcc), 2) parameters RMSE (ParamRMSE), 3) point set registration accuracy (PointAcc), 4) point set registration RMSE (PointRMSE), 5) Runtime (Runtime), and 6) optimization steps (OptiStep).
The parameters accuracy can be calculated as follows:

The parameters accuracy can be calculated as follows:
\[
Acc_{pm}  = {{{\rm{num}}({\rm{|}}{\bf{\hat \theta }} - {\bf{\theta }}^ *  {\rm{|}} < t_{pm} )} \mathord{\left/
		{\vphantom {{{\rm{num}}({\rm{|}}{\bf{\hat \theta }} - {\bf{\theta }}^ *  {\rm{|}} < t_{pm} )} {{\rm{num}}({\bf{\theta }}^ *  )}}} \right.
		\kern-\nulldelimiterspace} {{\rm{num}}({\bf{\theta }}^ *  )}}
\]
Here, ${\rm{num}}( * )$ means the numbers when expression $*$ is true. $t_{pm}$ is the threshold of parameter estimation. Here we set it at 1/22, because the potential initialized parameter scale is 22 ($S - (b + 1) = 22$ ), and IRON’s output is normalized to 1.

The parameters RMSE can be calculated as follows: 
\[
RMSE_{pm}  = ({{\sum\nolimits_{l = 1}^L {\left\| {\theta _l^ *   - \hat \theta _l } \right\|} _2^2 } \mathord{\left/
		{\vphantom {{\sum\nolimits_{l = 1}^L {\left\| {\theta _l^ *   - \hat \theta _l } \right\|} _2^2 } L}} \right.
		\kern-\nulldelimiterspace} L})^{{1 \mathord{\left/
			{\vphantom {1 2}} \right.
			\kern-\nulldelimiterspace} 2}} 
\]
Here, $L$ is the number of parameters to be optimized. 

Point set registration accuracy can be calculated as follows:
\[
Acc_{pt}  = {{{\rm{num(|}}T({\bf{u}},{\bf{\theta }}^ *  ) - {\bf{v}}{\rm{|}} < t_{pt} {\rm{)}}} \mathord{\left/
		{\vphantom {{{\rm{num(|}}T({\bf{u}},{\bf{\theta }}^ *  ) - {\bf{v}}{\rm{|}} < t_{pt} {\rm{)}}} {{\rm{num}}({\bf{v}})}}} \right.
		\kern-\nulldelimiterspace} {{\rm{num}}({\bf{v}})}}
\]
Here, we set $t_{pt}  = 2$ pixels.

Point set registration RMSE can be calculated as follows:
\[
RMSE_{pt}  = ({{\sum\nolimits_{n = 1}^{N_c } {\left\| {T({\bf{u}}_n ,{\bf{\hat \theta }}) - {\bf{v}}_n } \right\|_2^2 } } \mathord{\left/
		{\vphantom {{\sum\nolimits_{n = 1}^{N_c } {\left\| {T({\bf{u}}_n ,{\bf{\hat \theta }}) - {\bf{v}}_n } \right\|_2^2 } } {N_c }}} \right.
		\kern-\nulldelimiterspace} {N_c }})^{{1 \mathord{\left/
			{\vphantom {1 2}} \right.
			\kern-\nulldelimiterspace} 2}} 
\]
Here $N_c$ is the number of correlation point pairs.

\subsection{Experimental Results and Analysis}
\begin{figure}[htpb]
	\centering
	\subfloat[parameters accuracy]{\includegraphics[width=0.45\linewidth]{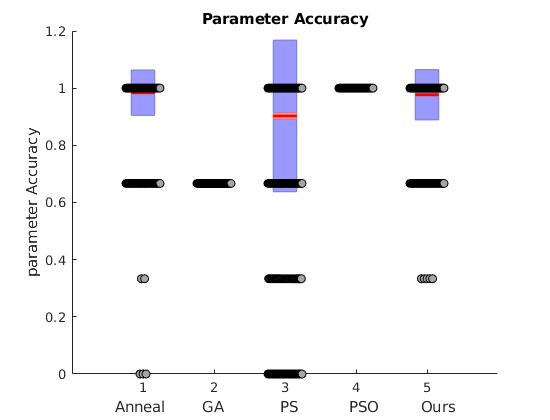}}
	\label{subfig:Uni_33}
	\hspace{1em}
	\subfloat[parameters RMSE]{\includegraphics[width=0.45\linewidth]{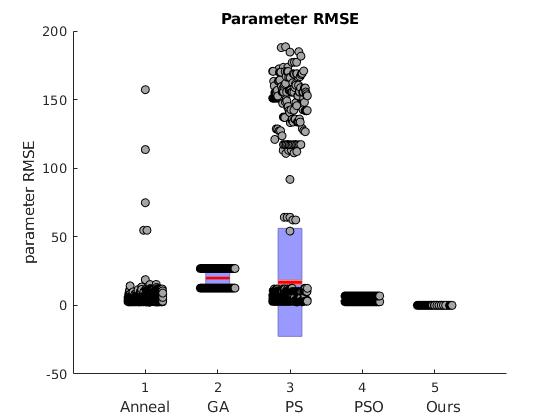}}
	\label{subfig:Uni_44}

	\subfloat[point registration accuracy]{\includegraphics[width=0.45\linewidth]{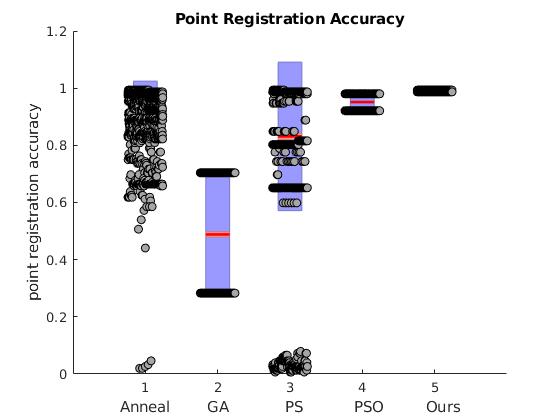}}
	\label{subfig:Uni_55}
	\hspace{1em}
	\subfloat[point registration RMSE]{\includegraphics[width=0.45\linewidth]{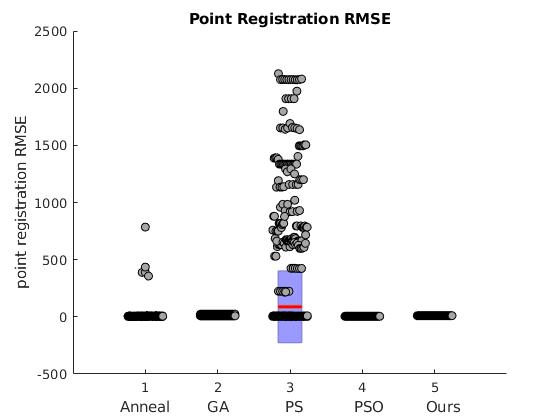}}
	\label{subfig:Uni_66}

	\subfloat[time]{\includegraphics[width=0.45\linewidth]{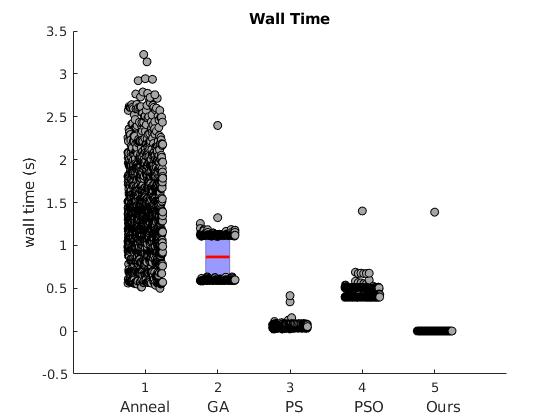}}
	\label{subfig:Uni_77}
	\hspace{1em}
	\subfloat[optimization steps]{\includegraphics[width=0.45\linewidth]{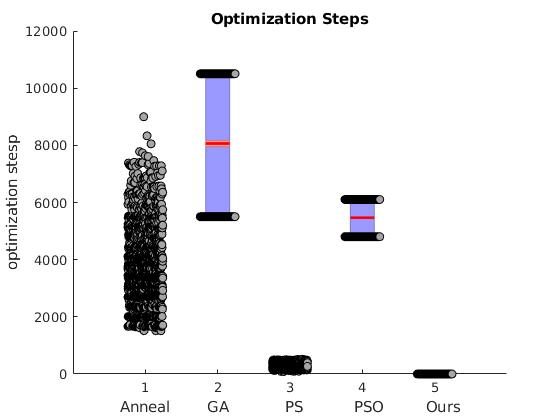}}
	\label{subfig:Uni_88}

	\caption{Uni-modal image registration results errorbars.}
	\label{fig:Unimodel-result}
\end{figure}

\begin{figure}[htpb]
	\centering
	\subfloat[parameters accuracy]{\includegraphics[width=0.45\linewidth]{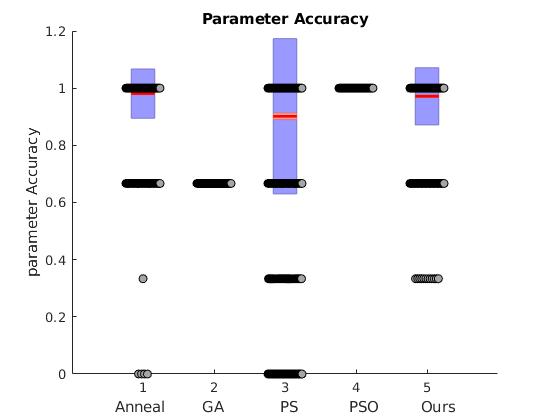}}
	\label{subfig:Multi_33}
	\hspace{1em}
	\subfloat[parameters RMSE]{\includegraphics[width=0.45\linewidth]{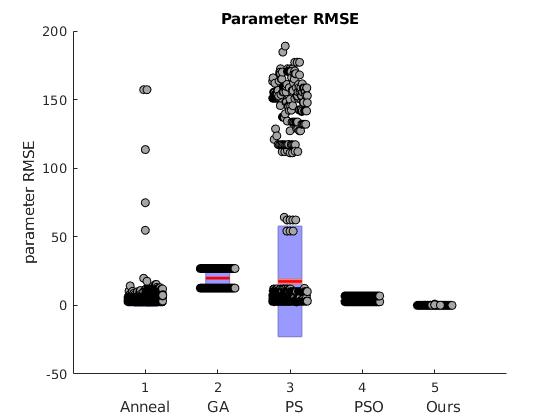}}
	\label{subfig:Multi_44}
	
	\subfloat[point registration accuracy]{\includegraphics[width=0.45\linewidth]{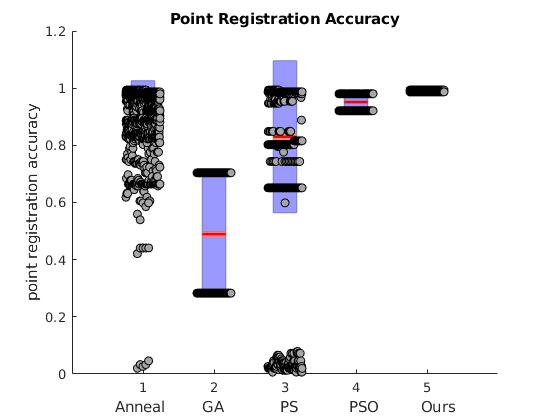}}
	\label{subfig:Multi_55}
	\hspace{1em}
	\subfloat[point registration RMSE]{\includegraphics[width=0.45\linewidth]{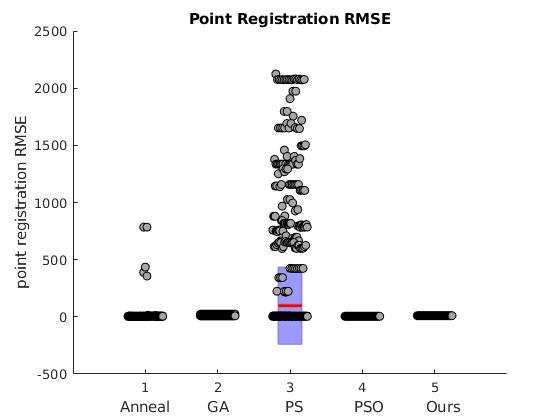}}
	\label{subfig:Multi_66}
	
	\subfloat[time]{\includegraphics[width=0.45\linewidth]{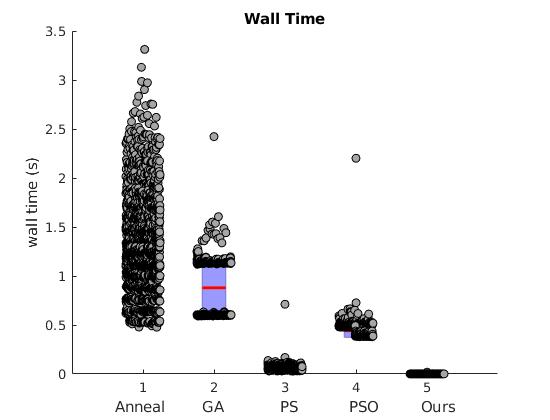}}
	\label{subfig:Multi_77}
	\hspace{1em}
	\subfloat[optimization steps]{\includegraphics[width=0.45\linewidth]{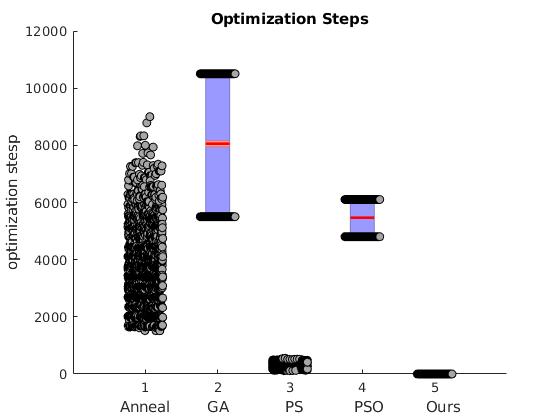}}
	\label{subfig:Multi_88}
	
	\caption{Multi-model image registration results errorbars.}
	\label{fig:Multimodel-result}
\end{figure}

For the uni-modal  image registration experiment, we employed a SIFT descriptor. The experimental results are shown in Fig.\ref{fig:Unimodel-result} and Table.\ref{tab:Uni_table}. For the multi-modal, we employed the Sobel operator to obtain identifying features. The results are shown in Fig.\ref{fig:Multimodel-result} and Table.\ref{tab:Multi_table}. Fig.\ref{fig:Unimodel-result} and Fig.\ref{fig:Multimodel-result} show the comparison by errorbars with notBoxPlot tools\footnote{https://uk.mathworks.com/matlabcentral/fileexchange/26508-notboxplot}. Table.\ref{tab:Uni_table} and Table.\ref{tab:Multi_table} list the statistical index of each algorithm’s performance, and the best performances are highlighted in bold. Both Fig.\ref{fig:Unimodel-result}(a)-\ref{fig:Unimodel-result}(f) and Fig.\ref{fig:Multimodel-result}(a)-\ref{fig:Multimodel-result}(f) show the errorbars of: 1) parameter accuracy, 2) parameter RMSE, 3) point registration accuracy, 4) point registration RMSE, 5) Runtime, and 6) optimization steps, respectively. Each errorbar demonstrates the experimental results from different optimization algorithms. They are: 1) simulated annealing, 2) genetic algorithm, 3) pattern search, 4) particle swarm optimization, and 5) IRON.

\begin{table}[]
	\caption{PERFORMANCES OF OPTIMIZATION BY DIFFERENT APPROACHES IN UNI-MODAL IMAGE REGISTRATION}
	\label{tab:Uni_table}
	\begin{tabular}{|c|c|c|c|c|c|c|}
		\hline
		\multicolumn{2}{|c|}{\multirow{2}{*}{Criterion}} & \multicolumn{5}{c|}{Methods}  \\ \cline{3-7} 
		\multicolumn{2}{|c|}{}   & Anneal & GA  & PS  & PSO  & ours  \\ \hline
		\multirow{2}{*}{\begin{tabular}[c]{@{}c@{}}ParamAcc\end{tabular}}     & mean & 0.984  & 0.667  & 0.903   & \textbf{1.000} & 0.977          \\ \cline{2-7} 
		& std  & 0.079  & 0.000  & 0.265   & \textbf{0.000} & 0.088          \\ \hline
		\multirow{2}{*}{\begin{tabular}[c]{@{}c@{}}ParamRMSE\end{tabular}}    & mean & 5.576  & 19.900 & 16.792  & 4.720          & \textbf{0.024} \\ \cline{2-7} 
		& std  & 5.286  & 7.130  & 39.321  & 2.087          & \textbf{0.015} \\ \hline
		\multirow{2}{*}{\begin{tabular}[c]{@{}c@{}}PointAcc\end{tabular}}     & mean & 0.931  & 0.488  & 0.831   & 0.951          & \textbf{0.990} \\ \cline{2-7} 
		& std  & 0.094  & 0.210  & 0.260   & 0.030          & \textbf{0.003} \\ \hline
		\multirow{2}{*}{\begin{tabular}[c]{@{}c@{}}PointRMSE\end{tabular}} & mean & 5.728  & 15.790 & 87.809  & \textbf{4.541} & 10.114         \\ \cline{2-7} 
		& std  & 24.626 & 7.029  & 314.049 & 0.378          & \textbf{0.102} \\ \hline
		\multirow{2}{*}{\begin{tabular}[c]{@{}c@{}}Runtime\end{tabular}}                & mean & 1.356  & 0.866  & 0.060   & 0.454          & \textbf{0.002} \\ \cline{2-7} 
		& std  & 0.458  & 0.265  & 0.047   & 0.059          & \textbf{0.000} \\ \hline
		\multirow{2}{*}{\begin{tabular}[c]{@{}c@{}}OptiStep\end{tabular}}  & mean & 3784.7 & 8064.0 & 324.0   & 5469.6         & \textbf{1.0}   \\ \cline{2-7} 
		& std  & 1245.2 & 2499.2 & 90.6    & 649.8          & \textbf{0.0}   \\ \hline
	\end{tabular}
\end{table}

\begin{table}[]
	\caption{PERFORMANCES OF OPTIMIZATION BY DIFFERENT APPROACHES IN MULTI-MODAL IMAGE REGISTRATION}
	\label{tab:Multi_table}
	\begin{tabular}{|c|c|c|c|c|c|c|}
		\hline
		\multicolumn{2}{|c|}{\multirow{2}{*}{Criterion}}  & \multicolumn{5}{c|}{Methods}  \\ \cline{3-7} 
		\multicolumn{2}{|c|}{}   & Anneal & GA     & PS      & PSO            & ours           \\ \hline
		\multirow{2}{*}{\begin{tabular}[c]{@{}c@{}}ParamAcc\end{tabular}}     & mean & 0.981  & 0.667  & 0.902   & \textbf{1.000} & 0.972          \\ \cline{2-7} 
		& std  & 0.086  & 0.000  & 0.272   & \textbf{0.000} & 0.100          \\ \hline
		\multirow{2}{*}{\begin{tabular}[c]{@{}c@{}}ParamRMSE\end{tabular}}    & mean & 5.634  & 19.878 & 17.466  & 4.726          & \textbf{0.028} \\ \cline{2-7} 
		& std  & 6.213  & 7.131  & 40.285  & 2.087          & \textbf{0.029} \\ \hline
		\multirow{2}{*}{\begin{tabular}[c]{@{}c@{}}PointAcc\end{tabular}}     & mean & 0.929  & 0.489  & 0.83    & 0.951          & \textbf{0.990} \\ \cline{2-7} 
		& std  & 0.097  & 0.21   & 0.266   & 0.030          & \textbf{0.003} \\ \hline
		\multirow{2}{*}{\begin{tabular}[c]{@{}c@{}}PointRMSE\end{tabular}}    & mean & 5.944  & 15.769 & 97.477  & \textbf{4.542} & 10.114         \\ \cline{2-7} 
		& std  & 28.926 & 7.03   & 337.255 & 0.378          & \textbf{0.102} \\ \hline
		\multirow{2}{*}{\begin{tabular}[c]{@{}c@{}}Runtime\end{tabular}}       & mean & 1.299  & 0.879  & 0.067   & 0.441          & \textbf{0.001} \\ \cline{2-7} 
		& std  & 0.461  & 0.275  & 0.022   & 0.067          & \textbf{0.000} \\ \hline
		\multirow{2}{*}{\begin{tabular}[c]{@{}c@{}}OptiStep\end{tabular}}     & mean & 3778.3 & 8056.5 & 323.3   & 5467.6         & \textbf{1.0}   \\ \cline{2-7} 
		& std  & 1272.0 & 2499.4 & 90.4    & 649.8          & \textbf{0.0}   \\ \hline
	\end{tabular}
\end{table}

\begin{figure}[htpb]
	\centering
	\subfloat[optical image]{\includegraphics[width=0.27\linewidth,height=0.27\linewidth]{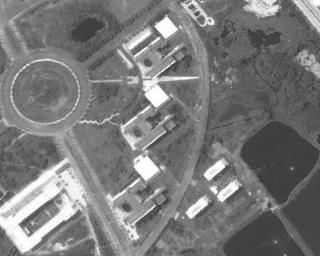}}
	\label{subfig:optical}
	\hspace{1em}
	\subfloat[reference image]{\includegraphics[width=0.27\linewidth,height=0.27\linewidth]{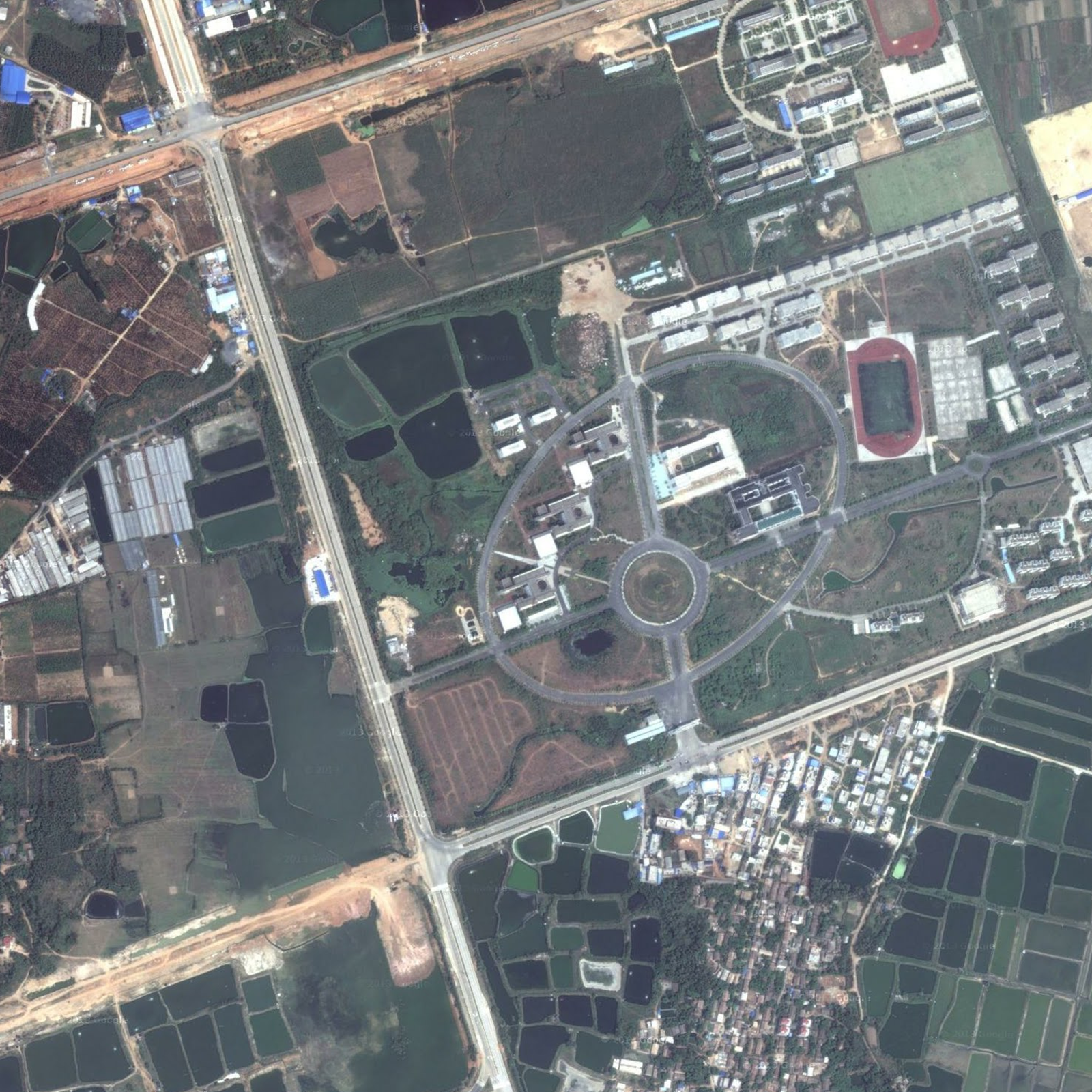}}
	\label{subfig:temp1}
	\hspace{1em}
	\subfloat[unimodel result]{\includegraphics[width=0.27\linewidth,height=0.27\linewidth]{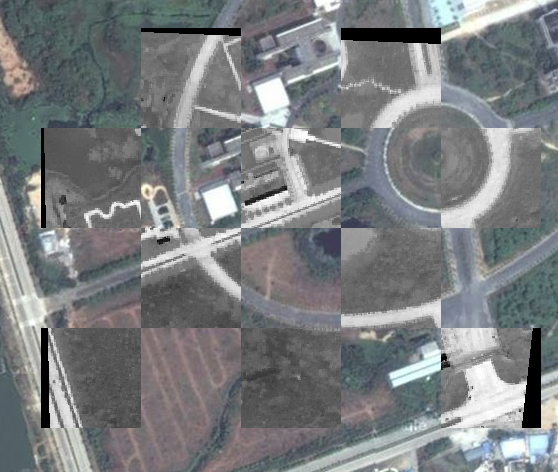}}
	\label{subfig:re_opt_mosiac}
	
	\subfloat[infrared image]{\includegraphics[width=0.27\linewidth,height=0.27\linewidth]{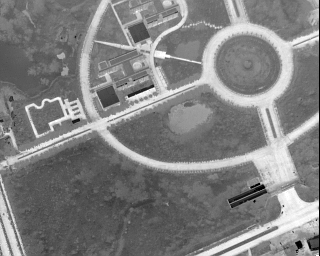}}
	\label{subfig:infrared}
	\hspace{1em}
	\subfloat[reference image]{\includegraphics[width=0.27\linewidth,height=0.27\linewidth]{re_goglmap-base.png}}
	\label{subfig:temp2}
	\hspace{1em}
	\subfloat[multimodel result]{\includegraphics[width=0.27\linewidth,height=0.27\linewidth]{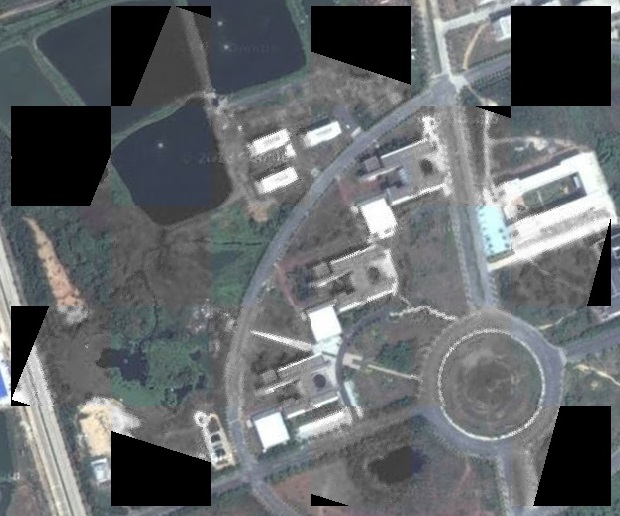}}
	\label{subfig:re_infra_mosiac}
	
	\captionsetup{justification=centering}
	\caption{Image registration mosaic results.}
	\label{fig:registration}
\end{figure}

From the experimental results, we find that our proposed approach executed the least optimization steps (only 1 step) and shortest runtime (0.001s) but achieved better performance. The PSO algorithm has a better parameter accuracy performance (1.000), but its parameter RMSE is worse, so that its point registration accuracy and point registration RMSE performance are worse. The experimental optical and infrared sensed image are shown in Fig.\ref{fig:registration}(a), \ref{fig:registration}(d), \ref{fig:registration}(b), and \ref{fig:registration}(e), and use an identical optical reference image. The uni-modal and multi-modal  image registration checkboards are shown in Fig.\ref{fig:registration}(c), and Fig.\ref{fig:registration}(f) respectively.

\section{Conclusion}
In this paper, for the purpose of alleviating the contradiction of the optimization algorithm in an image registration task, a learning-based image registration optimization network (IRON) was proposed. By trained with the similarity metric tensor whose label is a vector, the IRON acquired the structure of the search space, so that it could predict global optimal parameters immediately instead of iteratively searching. The experimental results demonstrate the superiority of our proposed approach, which performed better and with higher accuracy, lower RMSE, less runtime, and fewer optimization steps. Further work is needed to optimize more parameters by feeding higher order similarity metric tensors into an optimization network, which needs to implement an arbitrary dimension convolution function $convnd$.

\ifCLASSOPTIONcaptionsoff
  \newpage
\fi

%
%
%
%
%

\end{document}